\documentclass{article}
\usepackage[preprint]{neurips_2023}

\usepackage{amsmath}
\usepackage[utf8]{inputenc} 
\usepackage[T1]{fontenc}    
\usepackage{hyperref}       
\usepackage{url}            
\usepackage{booktabs}       
\usepackage{amsfonts}       
\usepackage{nicefrac}       
\usepackage{microtype}      
\usepackage{xcolor}         
\usepackage{graphicx}

\title{Graphical Reasoning: LLM-based Semi-Open Relation Extraction}

\author{
  Yicheng Tao\thanks{Equal contributions.} \\
  Department of EECS\\
  University of Michigan\\
  Ann Arbor, MI 48109 \\
  \texttt{yctao@umich.edu} \\
  \And
  Yiqun Wang$^*$ \\
  Department of DCMB \\
  University of Michigan\\
  Ann Arbor, MI 48109 \\
  \texttt{wyq@umich.edu} \\
  \And
  Longju Bai$^*$ \\
  Department of EECS \\
  University of Michigan\\
  Ann Arbor, MI 48109 \\
  \texttt{longju@umich.edu} \\
}

\begin{document}

\maketitle

\begin{abstract}
This paper presents a comprehensive exploration of relation extraction utilizing advanced language models, specifically Chain of Thought (CoT) and Graphical Reasoning (GRE) techniques. We demonstrate how leveraging in-context learning with GPT-3.5 can significantly enhance the extraction process, particularly through detailed example-based reasoning. Additionally, we introduce a novel graphical reasoning approach that dissects relation extraction into sequential sub-tasks, improving precision and adaptability in processing complex relational data. Our experiments, conducted on multiple datasets, including manually annotated data, show considerable improvements in performance metrics, underscoring the effectiveness of our methodologies.
\end{abstract}

\section{Introduction}

Relation extraction (RE) is a fundamental task in natural language processing (NLP) that involves identifying semantic relationships from text between specified entities. This task is pivotal in various applications such as building knowledge graphs, enhancing information retrieval, and powering recommendation systems. The complexity of accurately identifying these relationships from natural language has prompted the development of numerous approaches leveraging advancements in machine learning and, more recently, large language models (LLMs).

This paper introduces innovative methodologies employing advanced LLMs to enhance the accuracy and efficiency of relation extraction. We explore two specific approaches: Chain of Thought (CoT) with In-Context Learning, and Graphical Reasoning (GRE). These methodologies not only push forward the boundary of current RE capabilities but also introduce novel ways of structuring the RE process to handle increasingly complex data.

Chain of Thought with In-Context Learning utilizes the in-context learning capabilities of GPT-3.5, where the model is prompted with examples that illustrate a step-by-step reasoning process for extracting relationships from text. This approach mimics human problem-solving behavior, making the model's predictions more interpretable and reliable. We demonstrate this method's effectiveness through detailed examples and extensive testing across multiple datasets.

We proposed Graphical Reasoning for Relation Extraction as a new approach that decomposes the relation extraction task into sequential sub-tasks: entity recognition, text paraphrasing using recognized entities, and relation extraction based on the paraphrased text. This modular approach allows for targeted optimizations and improvements in each sub-task, leading to overall better performance. The graphical reasoning approach is detailed with a mathematical formulation, highlighting its theoretical foundations and practical implementations.

Empirical Evaluation is conducted to validate the effectiveness of our proposed methods. We perform experiments on several well-known datasets, including ADE, CoNLL04, and NYT. Additionally, we introduce a manually annotated version of the CoNLL04 dataset to address issues found in the original annotations and provide a more reliable testbed for our experiments.

The results of this project illustrate significant improvements in relation extraction capabilities using our proposed methods compared to traditional approaches. We provide a comprehensive analysis of the results, discuss the performance improvements brought by manual annotations, and demonstrate how each method contributes uniquely to the advancements in RE.

\textbf{Contributions:} The primary contributions of our work are the development of the Chain of Thought with In-Context Learning and Graphical Reasoning methodologies, empirical validation of these methods on standard datasets, and the enhancement of dataset quality through manual annotations. Additionally, we release our annotated dataset to the community to foster further research.

The introduction of these advanced methodologies in relation extraction showcases the potential of integrating structured reasoning and detailed problem decomposition in improving the performance and reliability of NLP tasks. This paper not only details our approaches and their validation but also sets the stage for future advancements in the application of LLMs in complex NLP tasks.








\section{Related Work}

\subsection{Relation Extraction with LLMs}
Given a sentence, the relation extraction task is to identify entities and their semantic relationships from texts. Closed, semi-open, and open relation extraction (RE) tasks utilize different strategies across supervised and unsupervised methods. 

In closed RE, fully-supervised methods such as KnowPrompt \cite{chen2022knowprompt} implement constraints between virtual node and relation type embeddings based on type semantics knowledge. PTR \cite{han2022ptr} uses prompt-tuning with Bert-based large language models (LLMs), masking type tokens for effective prompt adjustments. Additionally, a zero-shot approach by \cite{li2023revisiting} involves summarizing texts with GPT-3, generating and answering questions about relation validity based on these summaries to identify the most certain valid relations. 

In semi-open RE, \cite{cabot2021rebel} introduces a novel triplet linearization template and fine-tunes generative LLMs on these triplets, while \cite{wadhwa2023revisiting} employs Chain of Thoughts (CoT) explanations of training samples to fine-tune a smaller T5-based model, achieving state-of-the-art results in fully-supervised settings but falling short in few-shot scenarios. 

For open RE, methods include prompting generative LLMs like the LLaMA Family \citep{touvron2023llama} and the GPT Family \citep{brown2020language}, demonstrating the flexibility of generative models in handling open-ended tasks.

\subsection{Prompt Engineering for LLMs}
In recent years, few-shot in-context learning has emerged as a pivotal area of research, particularly in the realm of natural language processing. This learning paradigm leverages pre-trained language models to generalize from a limited number of examples, effectively learning new tasks without extensive retraining. A seminal paper by \cite{brown2020language} demonstrated that large language models like GPT-3 could perform a variety of tasks in a few-shot manner by simply providing them with a prompt that includes a few examples of the task at hand. This approach has been further explored in various studies, such as \cite{tam2021improving}, which investigates strategies for selecting optimal examples to include in the prompt, enhancing the model's performance on the target task. These developments underscore the potential of few-shot learning to adapt large-scale models to specific applications with minimal data, making it a promising direction for resource-constrained scenarios.

The concepts of Chain of Thought (CoT) and Tree of Thought (ToT) represent advanced strategies for tackling complex reasoning tasks. The CoT method, introduced by \cite{wei2023chainofthought}, involves prompting models to generate intermediate reasoning steps before arriving at a final answer, thereby mimicking human-like problem-solving processes. This approach has shown significant improvements in the model's ability to handle multi-step reasoning questions. Building on this, the Tree of Thought (ToT) framework extends the idea by structuring the reasoning process in a more hierarchical and branching manner, as discussed in \cite{yao2023tree}. This allows the model to explore multiple reasoning pathways and aggregate these to formulate a more robust conclusion. These methodologies not only enhance the interpretability of model decisions but also contribute to their robustness and accuracy in complex reasoning domains.

\section{Method}

\subsection{Chain of Thought with In-Context Learning}

The Chain of Thought with In-Context Learning approach leverages the inherent capabilities of GPT-3.5 to perform relation extraction through a few-shot learning framework. This method strategically utilizes in-context learning by embedding a set of 13 carefully curated examples into the model's prompt. Each example includes a text, a step-by-step explanation of the thought process (Chain of Thought), and the resultant relation triples, which guide the model in identifying and reasoning about relationships within given texts.

This approach is grounded in the idea that providing a model with structured examples that demonstrate both the problem-solving process and the final answer can enhance its ability to infer and generalize from limited data. The examples in the prompt cover various relation types such as 'Work For', 'Located In', 'Live In', and 'Kill', among others, and include entities classified as PERSON, LOCATION, ORGANIZATION, and OTHER. Each example begins with an instructional prefix that sets the task framework and is followed by the specific content and its analysis.

For instance, consider the following example included in the model's prompt:

\textbf{Instructional Prefix:} \textit{List the relations of the types [OrgBased In, Work For, Located In, Live In, Kill] among the entities [PERSON, LOCATION, ORGANIZATION, OTHER] in the given text and provide a reasonable explanation.}

\textbf{TEXT:} "\textit{Edward Marks, an official with the Montgomery County Democratic Party, argued that if Ms. Toth is not interested in the job, `she should get out...}"

\textbf{Explanation:} \textit{Edward Marks is an official that works for the Montgomery County Democratic Party.}

\textbf{Relations:} \textit{[["Edward Marks:Per", "Work For", "Montgomery County Democratic Party:Org"]]}

In this example, the model is prompted to dissect the text and generate an explanation leading to the identification of relationships. This step-by-step elucidation, modeled after the Chain of Thought approach, assists the model in not only parsing the textual information but also in logically connecting entities and their relationships, based on the organizational context provided. The use of a detailed explanatory step ensures that the reasoning behind each identified relation is both transparent and justifiable, increasing the reliability and robustness of the outcomes.

By integrating this method, we aim to enhance the model's capability to execute relation extraction tasks with high precision, fostering its ability to handle complex NLP challenges with minimal direct supervision and reliance on extensive training datasets. This combination of in-context learning and reasoned output illustrates the potent capabilities of advanced language models in addressing intricate tasks in natural language understanding.

\subsection{Graphical Reasoning}
Based on the idea of sub-tasking in Graph of Tasks, we divided the task of relation extraction into three sequential sub-tasks: (1) entity extraction, (2) text paraphrase using extracted entities, and (3) relation extraction. We first give a mathematical formulation of our method. Suppose the given entity types are $\mathcal{E}$, given relation types are $\mathcal{R}$, and the distribution of the text $\mathcal{T}$. We want to get 
$$
\{\langle \hat{e}_{s_i}, \hat{r}_i, \hat{e}_{o_i}\rangle\}_{i=1}^{k_t}=\mathop{argtop\_{k_t}}_{e_s, e_o\in \mathcal{E}, r\in \mathcal{R}, t\sim\mathcal{T}} P(\langle e_s, r, e_o\rangle\mid t)=\mathop{argmax}_{\langle e_{s_i}, r_i, e_{o_i}\rangle}\Pi_{i=1}^{k_t} P(\langle e_{s_i}, r_i, e_{o_i}\rangle\mid t)
$$
where we can further write
\begin{align*} 
\Pi_{i=1}^{k_t} P(\langle e_{s_i}, r_i, e_{o_i}\rangle\mid t) 
&= \Pi_{i=1}^{k_t} P(\langle e_{s_i}, r_i, e_{o_i}\rangle\mid t, \langle e_{s_i}, e_{o_i}\rangle)\cdot \Pi_{i=1}^{k_t} P(\langle e_{s_i}, e_{o_i}\rangle\mid t) \\ 
&= \Pi_{i=1}^{k_t} P(\langle e_{s_i}, r_i, e_{o_i}\rangle\mid \tilde{t}_{\langle e_s, e_o\rangle}, \langle e_{s_i}, e_{o_i}\rangle)\cdot P(e_{s_i}, \ldots e_{s_{k_t}}, e_{o_i}, \ldots e_{o_{k_t}}\mid t)
\end{align*}

where $\tilde{t}_{\langle e_s, e_o\rangle}$ means the paraphrased text of the original text $t$ taking in the entity information $\langle e_s, e_o\rangle$ we care about. For our case, we actually use all extracted entities to paraphrase the text for saving money. Since we are using ChatGPT that can only produce zero or one probability (yes or no) and can not afford the cost of running ChatGPT multiple times, all the probabilities in the above equations are zero or one. In the following sections, we will discuss how we design the prompt to get the result from the ChatGPT.

\subsubsection{Entity Extraction}
We use ChatGPT to detect entities within the given entity types. We find that adding precise scope of the entity types in the instruction prefix benefits ChatGPT. For example, in CONLL04, we use the instruction prefix: \emph{List the entities in [Per, Loc, Org] in the given text. Per only includes human names. Loc only includes location names shown on map, such as city, state, province, country, and country union. Org includes places other than location names shown on map, such as city, state, province, country, and country union.} For CONLL04, we use 13 examples as used in \cite{wadhwa2023revisiting}, e.g., \emph{TEXT: Rome is in Lazio province and Naples in Campania. Entities: ["Rome:Loc", "Lazio:Loc", "Naples:Loc", "Campania:Loc"]}. In the prompt, we first list instruction prefix, then examples, then instruction prefix, and finally sample with text we want to extract relations from, following the format of \cite{wadhwa2023revisiting}. For the other two sub-tasks, we also stick to this format.

\subsubsection{Text Paraphrase}
Inspired by \cite{li2023revisiting} and generative power of ChatGPT, we paraphrase the text using the extracted entities with prompt: \emph{Given the text: <TEXT>. Use the entities <ENTITIES> to paraphrase the text.} The instruction prefix is: \emph{Paraphrase the given text using the given entities.} By doing this, we actually find the text becomes more understandable, even for human readers. For example, 

\emph{The eight-pound bomb had a detonator charge, similar to a shotgun shell, that emits smoke when it hits the ground, said Bert Byers, spokesman for Cecil Field Naval Air Station.} 

is paraphrased to 

\emph{Bert Byers, spokesman for Cecil Field Naval Air Station, stated that the eight-pound bomb had a detonator charge resembling a shotgun shell, which releases smoke upon impact.} 

using extracted entities \emph{Bert Byers:Per} and \emph{Cecil Field Naval Air Station:Org}.

\subsubsection{Relation Extraction}
We first formulate candidate relations using all possible pairs of extracted entities with a valid relation, e.g., relation ``live in'' can only have a subject entity type ``Person'' and an object entity type ``Location''. Then we ask ChatGPT whether this candidate relation is valid given the paraphrased text to extract relations, e.g., \emph{Does(Did) Denver locate in Colorado correct? (Yes/No)}. Note in this sub-task, we do not use examples, so the inferring of ChatGPT will be faster than previous two tasks. So, even we need to check several candidate relations (max: about 30), the maximum running time of a sample is about 20 seconds.

\begin{figure*}[ht]
\centering
\includegraphics[width=\textwidth]{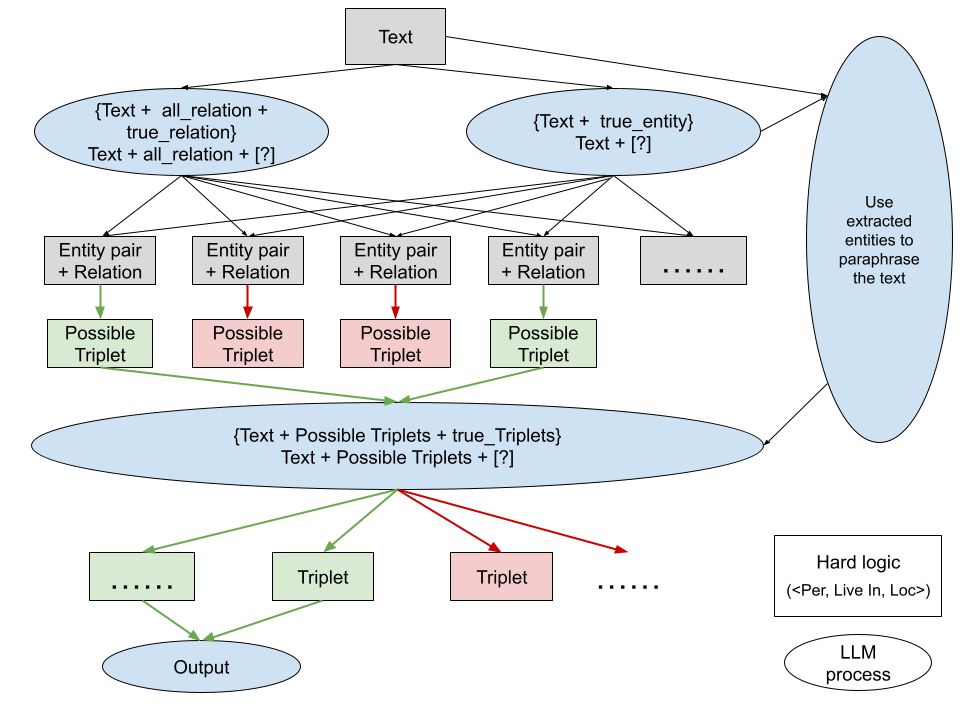}
\caption{Structure of GRE-GPT}
\label{fig:datasets-wide}
\end{figure*}

\section{Evaluation}
\subsection{Datasets}


For our relation extraction task, we utilized three well-known datasets, each offering unique characteristics and challenges. These datasets are ADE\citep{article}, CoNLL04\citep{Riedel2010ModelingRA}, and NYT\citep{Riedel2010ModelingRA}. Below, each dataset is described in detail, including the types of relations they include and the number of instances they contain.

\subsubsection{ADE (Adverse Drug Event)}

The ADE dataset is specifically designed for identifying adverse drug events from medical texts. This dataset is critical for applications within the healthcare sector, particularly for monitoring patient safety and for pharmacovigilance. The ADE dataset consists of annotated sentences where each sentence describes relations between drugs and their adverse effects. The primary relations are between drug names and disease symptoms, highlighting adverse drug reactions.

\textbf{Statistics:}
\begin{itemize}
    \item Relations included: Drug-Adverse Effect
    \item Number of instances: Approximately 4,272 sentences annotated with 6,800 drug-adverse effect relations.
\end{itemize}

\subsubsection{CoNLL04}

The CoNLL04 dataset is widely used in relation extraction tasks and serves as a benchmark for evaluating the performance of relation extraction models. It encompasses annotations for both entity and relation types, making it versatile for various NLP tasks. The dataset includes relations among entities like people, organizations, locations, and other miscellaneous entities.

\textbf{Statistics:}
\begin{itemize}
    \item Relations included: Person-Location, Organization-Person, Person-Person, etc.
    \item Number of instances: Contains about 1,400 sentences with detailed annotations of entities and relation types.
\end{itemize}

\subsubsection{NYT (New York Times)}

Derived from the New York Times Annotated Corpus, the NYT dataset is used primarily for relation extraction research and consists of a large number of news articles annotated with relation types. The relations are derived from Freebase and are linked to corresponding entities mentioned in the articles. This dataset is notable for its realistic and complex sentence structures, providing a robust challenge for relation extraction models.

\textbf{Statistics:}
\begin{itemize}
    \item Relations included: Multiple relation types including Person-Organization, Location-Organization, and others based on real-world scenarios.
    \item Number of instances: Over 1.18 million articles annotated with entity and relation types, although only a subset of these are commonly used in typical relation extraction tasks.
\end{itemize}

Each of these datasets presents unique challenges and opportunities for our Chain of Thought with In-Context Learning method, providing diverse contexts and scenarios to evaluate the effectiveness of our approach in relation extraction tasks.

\subsubsection{Manually Annotated Dataset (Complete CoNLL04)}

During our error analysis phase, we observed numerous inaccuracies in the original CoNLL04 dataset, primarily concerning incomplete or non-comprehensive relation triplets within the sentences. To address these issues, we conducted a thorough manual review of the instances in the dataset to verify their correctness. This review process involved checking each relation triplet for accuracy and exhaustiveness, ensuring that all potential relations within the texts were correctly identified and annotated.

This refined process resulted in the creation of a completed version of the CoNLL04 dataset, which we refer to as the Manually Annotated Dataset. This version is more comprehensive and accurate, containing annotations that are extensively checked and verified for completeness. 

\textbf{Impact of the Manually Annotated Dataset:}
Upon re-evaluating our models using this newly annotated dataset, we observed a noticeable improvement in performance metrics. Both the baseline performance and the performance of our Chain of Thought with In-Context Learning method showed significant enhancements. This improvement underscores the importance of dataset quality in achieving high accuracy in relation extraction tasks, and highlights the benefits of meticulous data verification and annotation refinement.

Additionally, we have released the refined dataset and the results of our experiments on our GitHub repository, allowing other researchers and practitioners to replicate our results, compare methodologies, or extend upon our work. The repository can be accessed at \url{https://github.com/LongjuBai/EECS598_LLM_RelationExtraction}.

The results of these experiments with the Manually Annotated Dataset not only provided insights into the potential of our method but also demonstrated the crucial role of accurate and comprehensive data in the development of effective NLP models.

\subsection{Evaluation Metrics}

The performance of our relation extraction method is evaluated using six distinct metrics: micro and macro versions of recall, precision, and the F1-score. These metrics provide a comprehensive assessment of the model's ability to correctly identify and classify relationship triplets within the text.

\subsubsection{Metric Calculation}

The evaluation process focuses on the following key aspects of metric calculation:

\begin{enumerate}
    \item \textbf{Micro Metrics:} These metrics aggregate the contributions of all classes to compute the average metric. In micro-average recall, precision, and F1-score, we calculate these metrics globally by considering the total true positives, false negatives, and false positives across all classes.
    
    \item \textbf{Macro Metrics:} Macro metrics, on the other hand, calculate the metric independently for each class and then take the average. This method treats all classes equally, regardless of their frequency in the dataset. Macro-average recall, precision, and F1-score provide a measure of how effectively the model performs on each class without being influenced by class imbalance.
    
    \item \textbf{Comparison and Analysis:} For each test instance, predicted relations are extracted and compared against the ground truth annotations. This comparison yields sets of true positives (correct predictions), false positives (incorrectly predicted relations that are not in the ground truth), and false negatives (missed relations that are in the ground truth but not predicted).
    
    \item \textbf{Accuracy Computation:} From these sets, we compute the recall, precision, and F1-score for each relation type. Recall measures the fraction of relevant instances that were retrieved, precision measures the fraction of retrieved instances that are relevant, and the F1-score is the harmonic mean of precision and recall, providing a balance between the two.
\end{enumerate}

This metric calculation process allows us to thoroughly evaluate the effectiveness of our relation extraction system, ensuring that it not only predicts relations accurately but also maintains consistency across different types of relations and datasets.


\subsection{Results Analysis}

The performance of two classifiers, Chain of Thought (CoT) and Graphical Reasoning (GRE), was evaluated on the CoNLL04 dataset both before and after manual annotation refinement. The evaluation metrics used were Micro Precision, Micro Recall, Micro F1, Macro Precision, Macro Recall, and Macro F1, providing a detailed assessment of each method's ability to accurately identify and classify relational data. The results are shown in Table~\ref{tab:extended-results}.

\begin{table}[h]
\small
\centering
\begin{tabular}{|l|c|c|c|c|c|c|}
\hline
\textbf{Classifier} & \textbf{Micro Prec} & \textbf{Micro Rec} & \textbf{Micro F1} & \textbf{Macro Prec} & \textbf{Macro Rec} & \textbf{Macro F1} \\
\hline
\multicolumn{7}{|c|}{\textit{CoNLL04 Dataset}} \\
\hline
CoT &0.3396 &0.5215 &0.3996 &0.2976 &0.4914 &0.3707 \\
GRE (ours) &\textbf{0.4364} &\textbf{0.5867} &\textbf{0.4941} &\textbf{0.3748} &\textbf{0.5700} &\textbf{0.4522} \\
\hline
\multicolumn{7}{|c|}{\textit{CoNLL04 Dataset (After Annotated)}} \\
\hline
CoT &0.4408 &0.5075 &0.4488 &0.3899 &0.4662 &0.4246 \\
GRE (ours) &\textbf{0.5862} &\textbf{0.6206} &\textbf{0.5985} &\textbf{0.5525} &\textbf{0.6085} &\textbf{0.5792} \\
\hline
\multicolumn{7}{|c|}{\textit{ADE Dataset}} \\
\hline
CoT &0.6505 &\textbf{0.6264} &0.6382 &0.6505 &\textbf{0.6264} &0.6382 \\
GRE (ours) &\textbf{0.7565} &0.6213 &\textbf{0.6822} &\textbf{0.7565} &0.6213 &\textbf{0.6822} \\
\hline
\end{tabular}
\caption{Comparison of Classifier Performance Across Datasets: This table presents the performance metrics for the Chain of Thought (CoT) and Graphical Reasoning (GRE) classifiers on the CoNLL04 dataset both before and after annotation, as well as performance metrics for CoT on the ADE dataset. Metrics include Micro and Macro Precision, Recall, and F1-Score, highlighting improvements and overall effectiveness.}

\label{tab:extended-results}
\end{table}

\subsubsection{Performance on Original CoNLL04 Dataset}

Initially, GRE outperformed the CoT approach across all metrics on the original CoNLL04 dataset. Specifically, GRE achieved a Micro Precision of 0.4364, Micro Recall of 0.5867, and Micro F1 of 0.4941, compared to CoT's Micro Precision of 0.3396, Micro Recall of 0.5215, and Micro F1 of 0.3996. Similarly, in Macro metrics, GRE demonstrated superior performance with a Macro F1 of 0.4522 versus CoT's 0.3707. These results suggest that GRE's method of leveraging graphical representations to reason about relationships offers a significant advantage in handling the complexities inherent in the CoNLL04 data.

\subsubsection{Performance on Annotated CoNLL04 Dataset}

With the manually annotated corrections to the CoNLL04 dataset, both methods showed improved performance, validating the impact of dataset quality on relation extraction tasks. GRE continued to exhibit superior performance, with increases in every metric: Micro Precision rose to 0.5862, Micro Recall to 0.6206, and Micro F1 to 0.5985. The Macro F1 also improved to 0.5792. Although the CoT method showed gains with a Micro Precision increase to 0.4408 and Micro F1 to 0.4488, GRE's improvements were more pronounced, indicating that the enhanced data quality disproportionately benefited the GRE approach, likely due to its robustness and ability to exploit more detailed relational cues in the data.

\subsubsection{Performance on ADE Dataset with Chain of Thought (CoT)}

The Chain of Thought (CoT) method's performance on the ADE dataset, which focuses on adverse drug events, demonstrated substantial effectiveness. Given that the dataset involves a single relation type, the micro and macro metrics are identical. For this dataset, CoT achieved a Micro and Macro Precision of 0.6505, a Micro and Macro Recall of 0.6264, and a Micro and Macro F1 of 0.6382.

\textit{Note on Ongoing Experiments:} The results for the Graphical Reasoning (GRE) method on the ADE dataset are still pending due to computational constraints and the time-consuming nature of our other experiments and annotation processes. Results for the GRE method on the NYT dataset are also underway. We plan to continuously update our findings on our GitHub repository as these results become available and as additional analyses are completed.

This approach ensures that the research community can access the most current data and results, fostering transparency and collaboration. Updates and additional data will be shared through our GitHub repository, which can be accessed at \url{https://github.com/LongjuBai/EECS598_LLM_RelationExtraction}.

\subsubsection{Summary}

These results underscore the effectiveness of the Graphical Reasoning method, particularly in contexts where relational complexity and data quality are critical. The significant improvement in both classifiers following the dataset annotation also highlights the importance of accurate and comprehensive training data in developing more effective NLP models. Future work will focus on further refining data annotation processes and exploring how enhancements in dataset quality can specifically benefit different extraction methods.

\section{Conclusion and Future Work}
In conclusion, this study illustrates the substantial benefits of integrating advanced prompting strategies and task decomposition in the field of relation extraction. Our Chain of Thought with In-Context Learning and Graphical Reasoning approaches have shown to not only enhance the performance of relation extraction tasks but also provide clearer pathways for reasoning, which are crucial for complex data interpretations.

\textbf{Future Work:}
\begin{itemize}
    \item \textit{Expansion of Data Annotation:} We plan to extend our manual annotation to additional datasets, improving the quality and size of data available for training RE models.
    \item \textit{Algorithmic Improvements:} Further refinements in the graphical reasoning approach will be explored, focusing on optimizing the sub-task processes and integrating more granular levels of relational complexity.
    \item \textit{Integration with Other NLP Tasks:} We aim to investigate the application of our proposed methods in other NLP tasks such as document summarization and question answering to evaluate their versatility and effectiveness in broader contexts.
    \item \textit{Real-world Applications:} Applying our RE techniques to real-world scenarios, such as legal document analysis and biomedical text processing, will be crucial in understanding their practical impact and areas for improvement.
\end{itemize}

Our continued efforts will focus on not only advancing the theoretical frameworks but also on implementing these innovations in practical applications, striving for broader impacts across various domains of natural language processing.

\newpage


\bibliographystyle{plainnat}
\bibliography{ref}

\begin{thebibliography}{12}
\providecommand{\natexlab}[1]{#1}
\providecommand{\url}[1]{\texttt{#1}}
\expandafter\ifx\csname urlstyle\endcsname\relax
  \providecommand{\doi}[1]{doi: #1}\else
  \providecommand{\doi}{doi: \begingroup \urlstyle{rm}\Url}\fi

\bibitem[Brown et~al.(2020)Brown, Mann, Ryder, Subbiah, Kaplan, Dhariwal, Neelakantan, Shyam, Sastry, Askell, et~al.]{brown2020language}
Tom Brown, Benjamin Mann, Nick Ryder, Melanie Subbiah, Jared~D Kaplan, Prafulla Dhariwal, Arvind Neelakantan, Pranav Shyam, Girish Sastry, Amanda Askell, et~al.
\newblock Language models are few-shot learners.
\newblock \emph{Advances in neural information processing systems}, 33:\penalty0 1877--1901, 2020.

\bibitem[Cabot and Navigli(2021)]{cabot2021rebel}
Pere-Llu{\'\i}s~Huguet Cabot and Roberto Navigli.
\newblock Rebel: Relation extraction by end-to-end language generation.
\newblock In \emph{Findings of the Association for Computational Linguistics: EMNLP 2021}, pages 2370--2381, 2021.

\bibitem[Chen et~al.(2022)Chen, Zhang, Xie, Deng, Yao, Tan, Huang, Si, and Chen]{chen2022knowprompt}
Xiang Chen, Ningyu Zhang, Xin Xie, Shumin Deng, Yunzhi Yao, Chuanqi Tan, Fei Huang, Luo Si, and Huajun Chen.
\newblock Knowprompt: Knowledge-aware prompt-tuning with synergistic optimization for relation extraction.
\newblock In \emph{Proceedings of the ACM Web conference 2022}, pages 2778--2788, 2022.

\bibitem[Gurulingappa et~al.(2012)Gurulingappa, Mateen, Roberts, Fluck, Hofmann-Apitius, and Toldo]{article}
Harsha Gurulingappa, Abdul Mateen, Angus Roberts, Juliane Fluck, Martin Hofmann-Apitius, and Luca Toldo.
\newblock Development of a benchmark corpus to support the automatic extraction of drug‐related adverse effects from medical case reports.
\newblock \emph{Journal of Biomedical Informatics}, http://dx.doi.org/10.1016/j.jbi.2012.04.008, 04 2012.
\newblock \doi{10.1016/j.jbi.2012.04.008}.

\bibitem[Han et~al.(2022)Han, Zhao, Ding, Liu, and Sun]{han2022ptr}
Xu~Han, Weilin Zhao, Ning Ding, Zhiyuan Liu, and Maosong Sun.
\newblock Ptr: Prompt tuning with rules for text classification.
\newblock \emph{AI Open}, 3:\penalty0 182--192, 2022.

\bibitem[Li et~al.(2023)Li, Wang, and Ke]{li2023revisiting}
Guozheng Li, Peng Wang, and Wenjun Ke.
\newblock Revisiting large language models as zero-shot relation extractors.
\newblock \emph{arXiv preprint arXiv:2310.05028}, 2023.

\bibitem[Riedel et~al.(2010)Riedel, Yao, and McCallum]{Riedel2010ModelingRA}
Sebastian Riedel, Limin Yao, and Andrew McCallum.
\newblock Modeling relations and their mentions without labeled text.
\newblock In \emph{ECML/PKDD}, 2010.
\newblock URL \url{https://api.semanticscholar.org/CorpusID:2386383}.

\bibitem[Tam et~al.(2021)Tam, Menon, Bansal, Srivastava, and Raffel]{tam2021improving}
Derek Tam, Rakesh~R Menon, Mohit Bansal, Shashank Srivastava, and Colin Raffel.
\newblock Improving and simplifying pattern exploiting training, 2021.

\bibitem[Touvron et~al.(2023)Touvron, Lavril, Izacard, Martinet, Lachaux, Lacroix, Rozi{\`e}re, Goyal, Hambro, Azhar, et~al.]{touvron2023llama}
Hugo Touvron, Thibaut Lavril, Gautier Izacard, Xavier Martinet, Marie-Anne Lachaux, Timoth{\'e}e Lacroix, Baptiste Rozi{\`e}re, Naman Goyal, Eric Hambro, Faisal Azhar, et~al.
\newblock Llama: Open and efficient foundation language models.
\newblock \emph{arXiv preprint arXiv:2302.13971}, 2023.

\bibitem[Wadhwa et~al.(2023)Wadhwa, Amir, and Wallace]{wadhwa2023revisiting}
Somin Wadhwa, Silvio Amir, and Byron~C Wallace.
\newblock Revisiting relation extraction in the era of large language models.
\newblock In \emph{Proceedings of the conference. Association for Computational Linguistics. Meeting}, volume 2023, page 15566. NIH Public Access, 2023.

\bibitem[Wei et~al.(2023)Wei, Wang, Schuurmans, Bosma, Ichter, Xia, Chi, Le, and Zhou]{wei2023chainofthought}
Jason Wei, Xuezhi Wang, Dale Schuurmans, Maarten Bosma, Brian Ichter, Fei Xia, Ed~Chi, Quoc Le, and Denny Zhou.
\newblock Chain-of-thought prompting elicits reasoning in large language models, 2023.

\bibitem[Yao et~al.(2023)Yao, Yu, Zhao, Shafran, Griffiths, Cao, and Narasimhan]{yao2023tree}
Shunyu Yao, Dian Yu, Jeffrey Zhao, Izhak Shafran, Thomas~L. Griffiths, Yuan Cao, and Karthik Narasimhan.
\newblock Tree of thoughts: Deliberate problem solving with large language models, 2023.

\end{thebibliography}

\end{document}